\documentclass[11pt,a4paper]{article}
\usepackage[hyperref]{acl2021}
\usepackage{times}
\usepackage{mathptmx} 

\usepackage{multirow}
\usepackage{latexsym}

\usepackage{flexisym}
\usepackage{graphicx}
\usepackage{microtype}
\RequirePackage[OT1]{fontenc}

\aclfinalcopy 
\def\aclpaperid{2823X-A3F9H3F6P6} 

\newcommand{\commentEx}[1]{\texttt{\fontsize{9pt}{9pt}\selectfont#1}}

\title{Improving Persian Relation Extraction Models by Data Augmentation}

\author{
	Moein Salimi Sartakhti \\
	Shahid Beheshti University \\
	Tehran, Iran \\
	\commentEx{sartakhti.salimi@gmail.com} \\ \And
	Romina Etezadi \\
  Shahid Beheshti University \\
  Tehran, Iran \\
 \commentEx{ro.etezadi@mail.sbu.ac.ir} \\ \And
  Mehrnoush Shamsfard \\
  Shahid Beheshti University \\
  Tehran, Iran \\
 \commentEx{m-shams@sbu.ac.ir} \\
}
\date{}

\begin{document}
\maketitle
\begin{abstract}
Relation extraction that is the task of predicting semantic relation type between entities in a sentence or document is an important task in natural language processing. Although there are many researches and datasets for English, Persian suffers from sufficient researches and comprehensive datasets. The only available Persian dataset for this task is PERLEX, which is a Persian expert-translated version of the SemEval-2010-Task-8 dataset. 
In this paper, we present our augmented dataset and the results and findings of our system, participated in the Persian relation Extraction shared task of NSURL 2021 workshop. We use PERLEX as the base dataset and enhance it by applying some text preprocessing steps and by increasing its size via data augmentation techniques to improve the generalization and robustness of applied models. We then employ two different models including ParsBERT and multilingual BERT for relation extraction on the augmented PERLEX dataset. Our best model obtained 64.67\% of Macro-F1 on the test phase of the contest and it achieved 83.68\% of Macro-F1 on the test set of PERLEX.

\end{abstract}

\begin{figure*}
	\centering
	\includegraphics[width=0.8\linewidth]{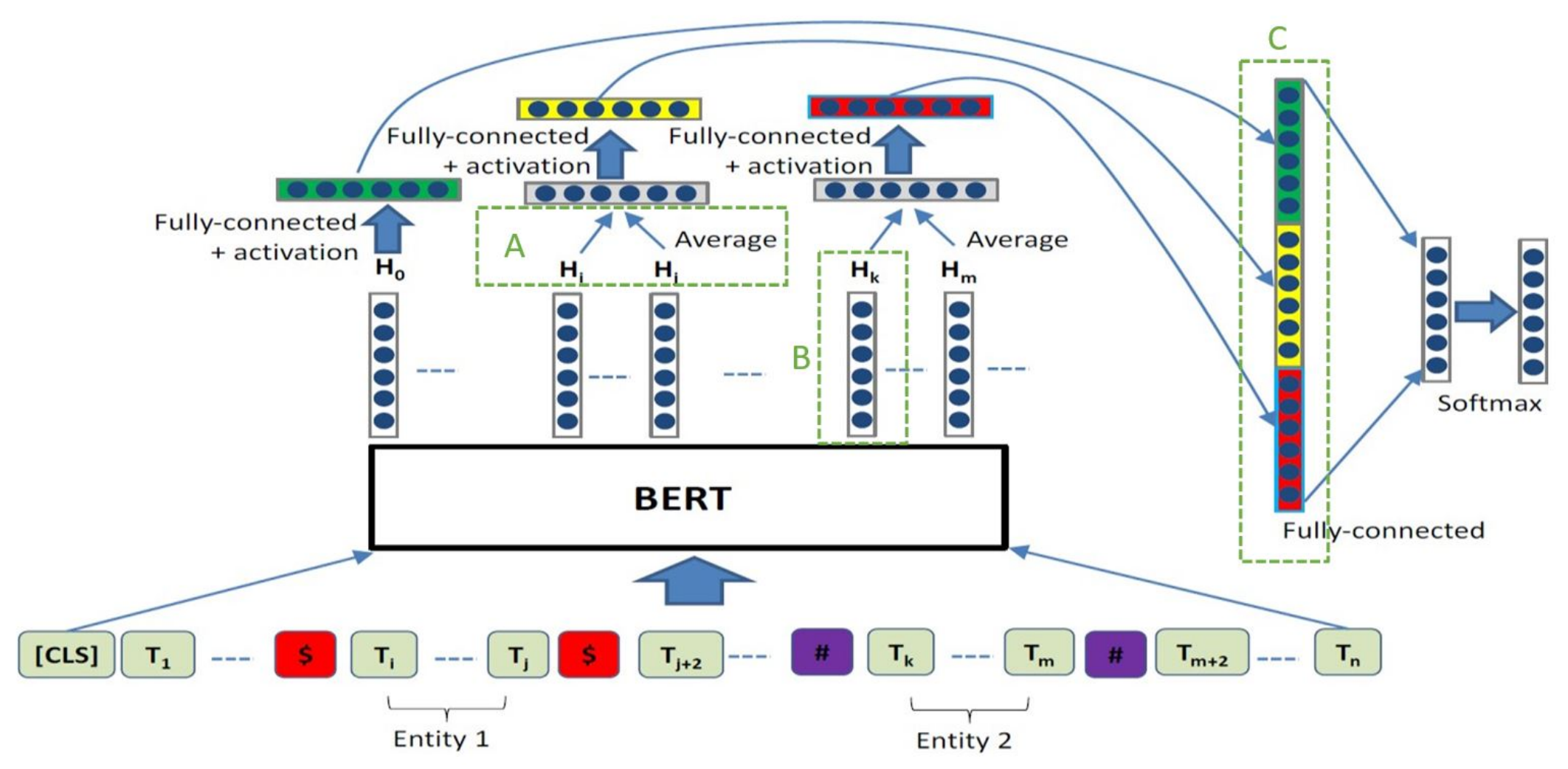}
	\caption{R-BERT structure.}
	\label{fig:rbert}
\end{figure*}

\section{Introduction}
The task of detecting semantic relations between entities in a text is called Relation Extraction (RE). RE plays an important role in various natural language processing (NLP) tasks such as Information Extraction, Knowledge Extraction, Question Answering, Text Summarization, etc.
According to the literature, RE tasks can be divided into two categories: sentence-level and document-level. The goal of the sentence-level RE task is to obtain the relation between two known entities (predefined entities) in a sentence. Nevertheless, the document-level RE task aims to extract the relationship among several entities in a long text which usually contains multiple sentences. According to the differences mentioned earlier, document level relation extraction is more complicated than sentence-level.

In the RE task, entities are string literals that are marked in the sentence and the aim is to identify a limited number of predefined relationships between these entities from the input text. Different tasks can benefit from using RE. For example, suppose that the goal of an information extraction system is to extract corporations located in Iran from a text. For this purpose, the RE component may use the \textit{located-in} predicate and \textit{Iran} as the object of the relation to allow this information to be extracted. Moreover, suppose a question answering system, which is going to answer a question about the cause of an event. It may exploit an RE task in which the relationship is \textit{Cause-Effect} and the object should be that specific event \citep{asgari-bidhendi2021perlex}.

Another important application of RE is knowledge base creation. A knowledge base includes a set of entities and relationships between them. Most of the available large knowledge bases such as Yago \citep{suchanek2007yago}, Freebase \citep{bollacker2008freebase}, DBpedia \citep{auer2007dbpedia}, and Wikidata \citep{vra2014wikidata} are encoded in English. In Persian, there is a knowledge base (knowledge graph) called Farsbase \citep{asgari-bidhendi2019farsbase}.
There are some standard RE datasets for the English language, such as SemEval-2010-Task 8 and TACRED. For Persian which is a low resource language in this field, the only RE dataset (up to authors' knowledge) is PERLEX, which is an expert-translated version of SemEval- 2010-Task-8 dataset.

PERLEX has 10717 sentences and there is a relation and two entities in each sentence. In PERLREX, the boundaries of each entity have been specified by certain tokens. For example, the first entity uses the tags \(<\)e1\(>\) and \(<\)/e1\(>\) for the start and end of the entity (also \(<\)e2\(>\) and \(<\)/e2\(>\) are used for the second entity). Table \ref{exmp} shows some examples of annotated sentences.

Our contributions in this work are as follows: (1) Using text augmentation techniques to increase the size of the PERLEX dataset. (2) Preprocessing the PERLEX to fix some of the issues which improves the performance of the latest Persian relation extractor.
In this paper, a relation extraction system is presented which is submitted to the Second Workshop on NLP Solutions for Under Resourced Languages (NSURL 2021). Some modifications on available models are adopted and the effects of each modification on the total generalization and robustness are reported.
The remainder of this paper is organized as follows: the methodology is described in Section 2. Section 3 shows the experimental results. Section 4 concludes the paper.

\begin{figure}
	\centering
	\includegraphics[width=0.9\linewidth]{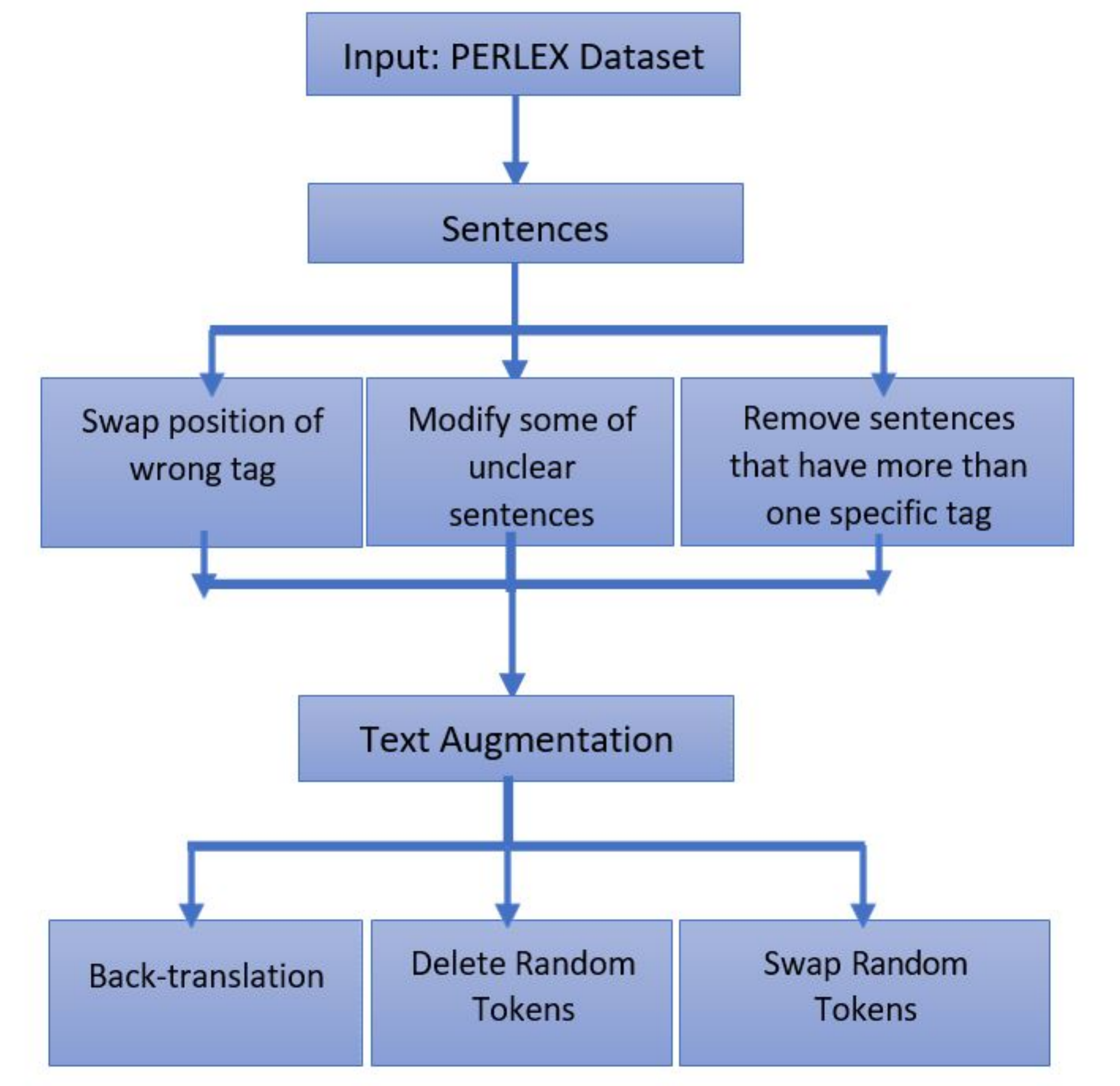}
	\caption{Text preprocessing and text augmentation procedure.}
	\label{fig:steps}
\end{figure}

\begin{figure*}
	\centering
	\includegraphics[width=0.8\linewidth]{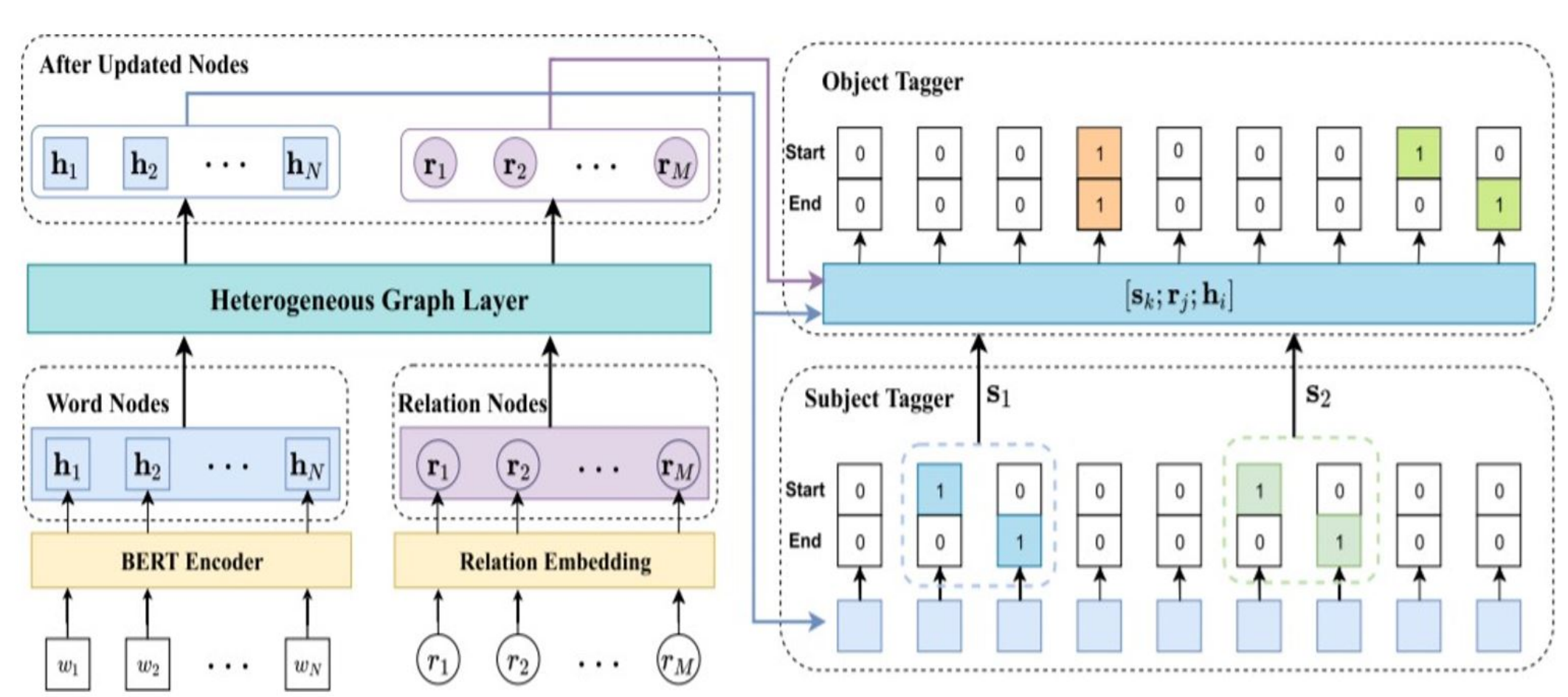}
	\caption{RIFRE structure.}
	\label{fig:rifire}
\end{figure*}

\begin{table*}
	\centering
	\caption{\label{exmp} Some correct and wrong examples of the PERLEX.}
	\begin{tabular}{ll}
		\hline \textbf{Relation Type} & \textbf{Sentence}  \\ \hline
		Product-Producer(e2,e1) & The \(<\)e1\(>\)company\(<\)/e1\(>\) fabricates plastic \(<\)e2\(>\)chairs\(<\)/e2\(>\).  \\
		
		\multicolumn{1}{l}{\multirow{2}{*}{Message-Topic(e1,e2)}} & The major theme of the \(<\)e1\(>\)book\(<\)/e1\(>\) is the \(<\)e2\(>\)beauty\(<\)/e2\(>\) \\
		\multicolumn{1}{l}{} & of a dream. \\
		
		Entity-Destination(e1,e2) & He has just sent \(<\)e1\(>\)spam\(<\)/e1\(>\) to the \(<\)e2\(>\)clients\(<\)/e2\(>\).  \\
		
		\multicolumn{1}{l}{\multirow{3}{*}{Message-Topic(e1,e2)}} & I read the \(<\)e1\(>\)report\(<\)/e1\(>\) from Somalia on the  \\
		\multicolumn{1}{l}{} & \(<\)e2\(>\)agreement\(<\)/e2\(>\) reached by faction leaders on the form of a future \\
		\multicolumn{1}{l}{} &  government that has \(<\)e2\(>\)been\(<\)/e2\(>\) warmly welcomed. \\
		
		\hline
	\end{tabular}
\end{table*}

\section{Methodology}
\subsection{Data Preprocessing}
Although there are many datasets for English and other rich-resource languages, Persian has no comprehensive available resources for the RE task. Data annotating is a challenging, time-consuming, and cost-consuming task. Therefore, in the data preprocessing step we try to leverage techniques like text augmentation to increase the size of PRELEX. Some preprocessing is also applied to PERLEX. The preprocessing and text augmentation steps are shown in Figure \ref{fig:steps}.

The preprocessing and text augmentation procedure both includes three sub-steps. Text preprocessing sub-steps are listed below:
\begin{itemize}
	\item Swap position of the wrong tag
	\item Modify the unclear sentences
	\item Remove sentences which have more than one specific tag
\end{itemize}
 
As PERLEX is translated semi-automatically, there are some problems in it, such as:
\begin{itemize}
	\item Some of the sentences have more than one tag \(<\)e1\(>\) or \(<\)/e1\(>\) or \(<\)e2\(>\) or \(<\)/e2\(>\). As it is supposed that each sentence contains one relation, such sentences are filtered. 975 sentences have this problem and are removed from the dataset (See the 4th sentence in Table \ref{exmp}).
	\item In all of the sentences that \(<\)e2\(>\) (\(<\)e1\(>\)) comes exactly before \(<\)/e1\(>\) (\(<\)/e2\(>\)), position of these tags is swapped. This issue is fixed by detecting these sentences and swapping the tokens. 344 sentences have this problem.
	\item Some of the unclear translated sentences in PERLEX have been modified.
\end{itemize}

After the data preprocessing step, some noise are added and the text augmentation techniques are applied to increase the size of the PERLEX. Some of the employed techniques are listed below:
\begin{itemize}
	\item Deleting a token in each sentence randomly
	\item Swapping positions of some tokens randomly
	\item Using the Back-translation method \citep{shleifer2019low} in order to increase the size of PERLEX dataset.
\end{itemize}

There are different ways for back-translating. For example, one way can be the translation of sentences to English, then to Arabic, and finally, return sentence to Persian. However, in this paper, each sentence is translated from Persian to English and then it is back-translated to Persian by using the python API of the google translate package\footnote{https://pypi.org/project/googletrans/}. Therefore, this method can increase PERLEX size from 9381 to 18762. Reaching 18762 sentences for Persian is an important achievement in the RE task.

\subsection{Applied Models}
This section describes different models that the data augmentation is applied on them: R-BERT \citep{wu2019enriching} and RIFRE \citep{zhao2021representation}. After the preprocessing and text augmentation steps, two state-of-the-art models R-BERT and RIFRE are employed.

\paragraph{R-BERT:} The main structure of R-BERT is shown in Figure \ref{fig:rbert}. For a sentence with two target entities e1 and e2, ‘\$’ has been inserted at both the beginning and end of the first entity, and ‘\#’ at both the beginning and end of the second entity. Also, there is a ‘[CLS]’ symbol at the beginning of each sentence. We finetune the pre-trained ParsBERT \citep{farahani2021parsbert} and Multilingual BERT \citep{libovi2019how} models on the augmented PERLEX. In addition, table \ref{dev1} shows other hyperparameters of R-BERT. Furthermore, we experiment with different combination of embeddings produced by R-BERT to reach the best model (See embeddings A, B,and C in Figure \ref{fig:rbert}).

Some of the modifications on the R-BERT are listed below:
\begin{itemize}
	\item R-BERT\_V1: Average all of the three final embeddings in the fully connected layer rather than a concatenation of them (see Figure \ref{fig:rbert}-C).
	\item R-BERT\_V2: Concatenation all of the embeddings of tokens in each entity rather than average them (Figure \ref{fig:rbert}-A).
	\item Using the last (first) token instead of average all of the embeddings of tokens in the entities (Figure \ref{fig:rbert}-B).
	\item Using the Multilingual BERT and ParsBERT to reach the best decision
\end{itemize}

\paragraph{RIFRE:} This work proposes a representation iterative fusion based on a heterogeneous graph neural network for joint entity and relation extraction. As shown in Figure \ref{fig:rifire}, RIFRE models relations and words as nodes on the graph and update the nodes through a message passing mechanism. The model performs relation extraction after nodes are updated. First, the subject tagger is used to detect all possible subjects on the word nodes. Then, RIFRE combines each word node with the candidate subject and relation, and the object tagger is used to tag the object on the new word nodes. In this paper, RIFRE is adopted with the ParsBERT and Multilingual BERT.

\begin{table}
	\centering
	\caption{\label{dev1} Parameters settings for the R-BERT model.}
	\begin{tabular}{ll}
		\hline \textbf{Parameters} & \textbf{Value}  \\ \hline
		Batch size & 16  \\
		Max sentence length & 128  \\
		Adam learning rate & 2e-5  \\
		Number of epochs & 10  \\
		Dropout rate & 0.1  \\
		\hline
	\end{tabular}
\end{table}

\begin{table}
	\centering
	\caption{\label{dev2} Parameters settings for the RIFRE model.}
	\begin{tabular}{ll}
		\hline \textbf{Parameters} & \textbf{Value}  \\ \hline
		Batch size & 16  \\
		Max sentence length & 128  \\
		Adam learning rate & 1e-1  \\
		Number of epochs & 10  \\
		Dropout rate & 0.1  \\
		\hline
	\end{tabular}
\end{table}

\begin{figure}
	\centering
	\includegraphics[width=\linewidth]{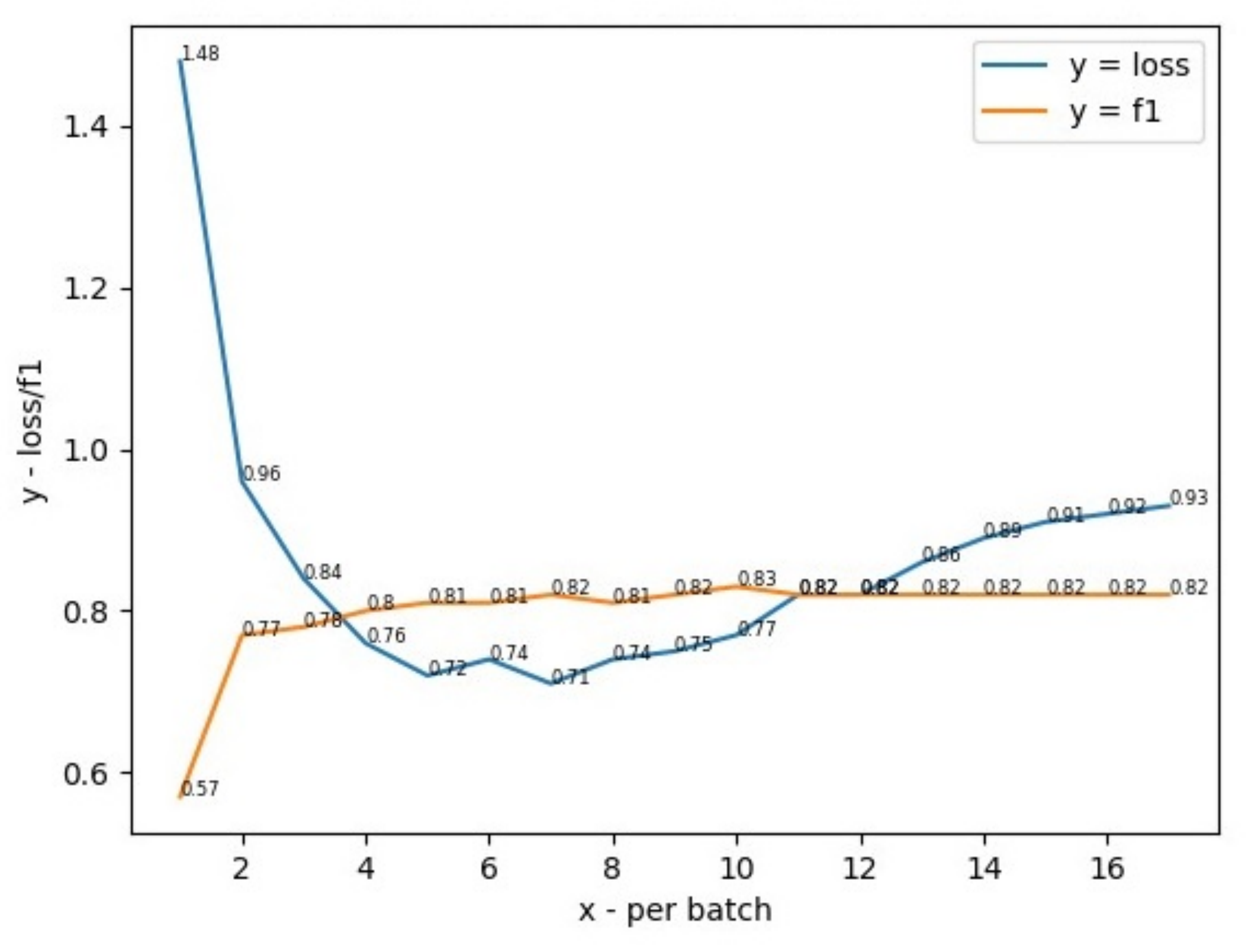}
	\caption{F-Score and Loss per epochs on the V1 R-BERT}
	\label{fig:nem1}
\end{figure}

\begin{figure}[t]
	\centering
	\includegraphics[width=\linewidth]{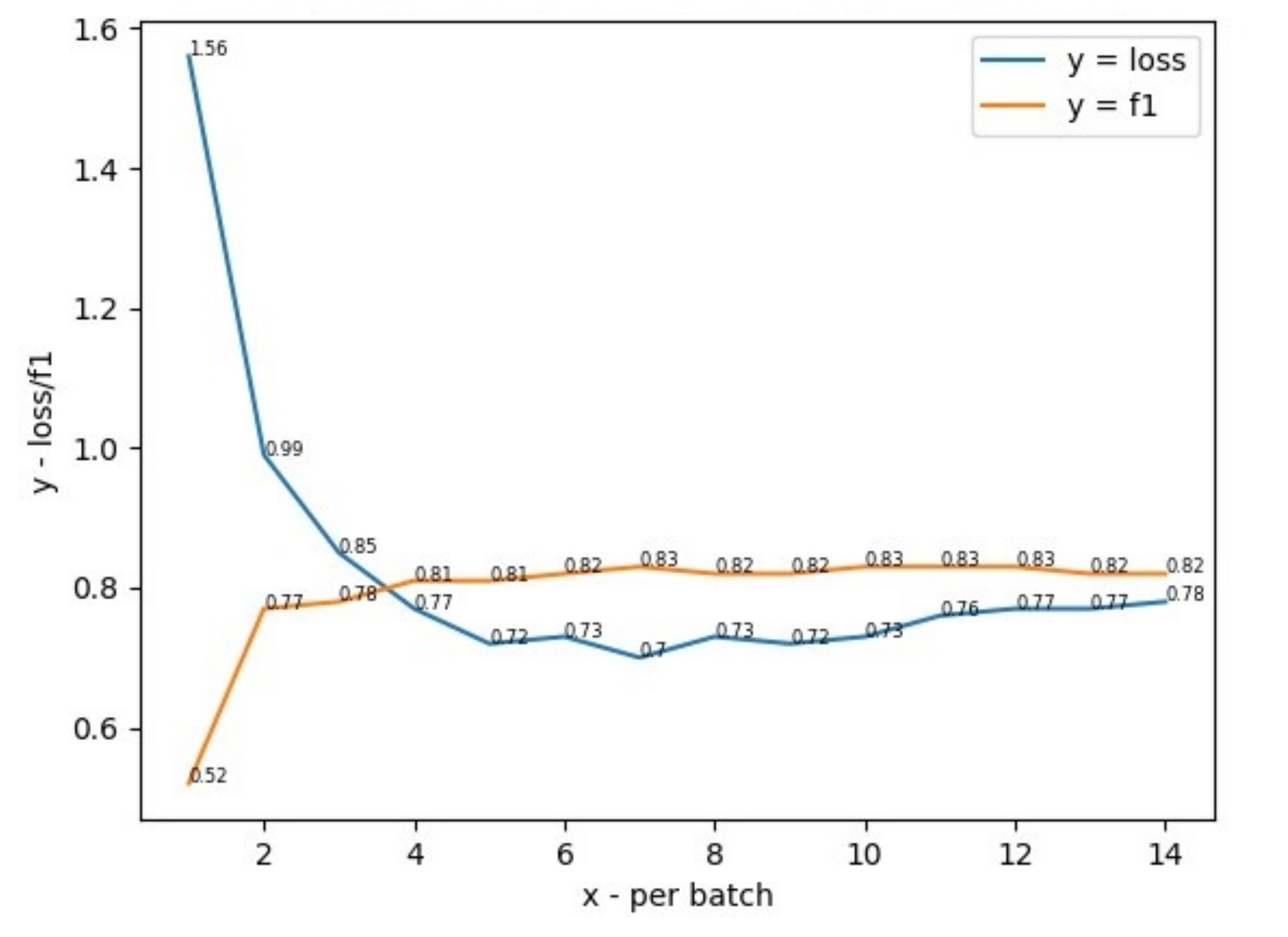}
	\caption{F-Score and Loss per epochs on the V2 R-BERT}
	\label{fig:nem2}
\end{figure}

\begin{figure}[t]
	\centering
	\includegraphics[width=\linewidth]{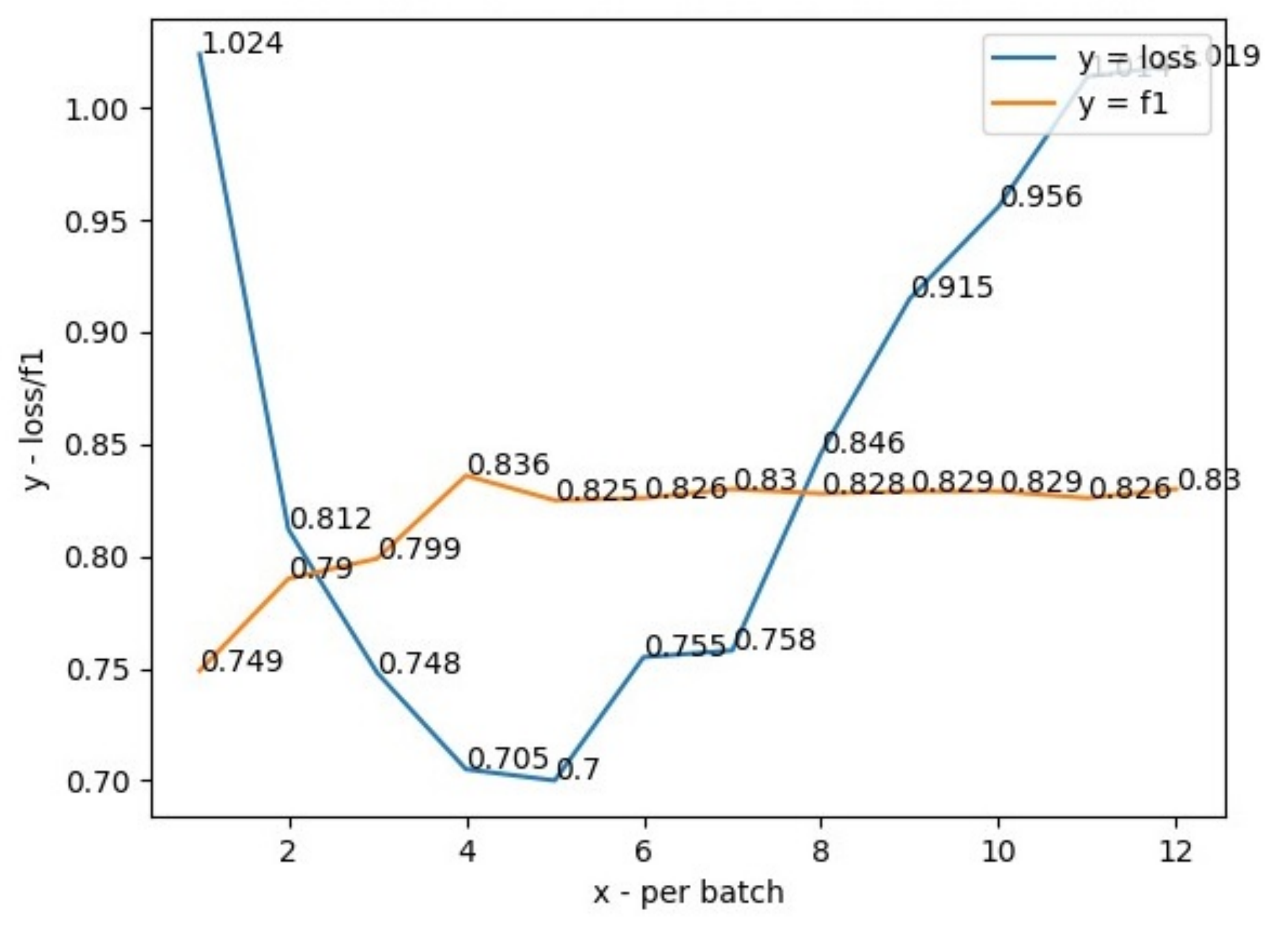}
	\caption{F-Score and Loss per epochs on the V3 R-BERT}
	\label{fig:nem3}
\end{figure}

\section{Evaluation}
There are three main ways to evaluate the RE classification results:
\begin{itemize}
	\item Taking into account both variations of each class (18 classes in total).
	\item Using only one variation of each class (and considering directionality).
	\item Using only one variation of each class (and ignoring directionality).
\end{itemize}

Moreover, there are two approaches to calculate F1-score: Micro-averaging and Macro-averaging. In this dataset, those pairs of entities that do not fall into any of the main nine classes are labeled as the "Other" class. The "Other" class is not participated in the evaluation phase. In this section, the official evaluation method is used for the SemEval-2010-Task-8 dataset, which is (9+1)-way classification with macro-averaging F1-score measurement while directionality is taken into account. This (9+1)-way means that the nine main classes plus “Other” in training and testing is considered, but "Other" is ignored to calculate the F1-scores.

%

\begin{table}[t]
	\centering
	\caption{\label{dev3} Performance of the models on PERLEX.}
	\begin{tabular}{ll}
		\hline \textbf{Models} & \textbf{F1-score}  \\ \hline
		Simple R-BERT & 83.86\%  \\
		R-BERT\_V1 & 83.02\%  \\
		R-BERT\_V2 & 83.11\%  \\
		R-BERT\_V3 & 83.08\%  \\
		RIFRE & 79.54\%  \\
		\hline
	\end{tabular}
\end{table}

\begin{table}
	\centering
	\caption{\label{dev4} Performance of the models on different relations types in PERLEX.}
	\begin{tabular}{ll}
		\hline \textbf{Relation Types} & \textbf{F1-score}  \\ \hline
		Cause-Effect &61.70\%\\
		Content-Container &59.26\%\\
		Entity-Destination &76.01\%\\
		Entity-Origin &58.04\%\\
		Instrument-Agency &75.54\%\\
		Member-Collection &32.85\%\\
		Message-Topic &76.06\%\\
		Other &40.95\%\\
		\hline
	\end{tabular}
\end{table}

\section{Results}
\subsection{Development Phase}
In the development phase, PERLEX dataset is used and some improvements are achieved. Table \ref{dev1} shows the major parameters used in R-BERT experiments. Hyperparameters of the RIFRE are shown in Table \ref{dev2}. Table \ref{dev3} shows the performance of the various models which are used. R-BERT model produces the best results, while RIFRE model produces the worst according to table \ref{dev3}. Figures \ref{fig:nem1}, \ref{fig:nem2} and \ref{fig:nem3} show the loss and F1-score value per epochs. According to these evaluations, simple R-BERT has better results than V1, V2, and V3 variation of the R-BERT. As table \ref{dev3} shows all of the results, the best model is the simple R-BERT which has achieved F1-Score 83.68 on the test set.

\subsection{Test Phase}
Finally, results show that the proposed model reaches 64.67 of Macro-F1 score on the shared task test data in NSURL contest.

%
%
%

\section{Conclusion}
In this paper, the PERLEX dataset is used which is a Persian expert-translated version of the "SemEval-2010-Task-8" dataset. As data annotating is a challenging, time-consuming and cost-consuming task, we employ some of the text preprocessing and text augmentation techniques such as back-translation, deleting random tokens, and swapping random tokens. The Preprocessing and text augmentation could increase F-Score by about four percent in comparison to the last and best work on Persian. After preparing the PERLEX, we apply two state-of-the-art models namely R-BERT and RIFRE. In addition, we extend the R-BERT model by changing the R-BERT structure. Pre-trained BERT models that are tested in this paper are ParsBERT and Multilingual Bert. Results show that ParsBERT based on the simple R-BERT structure had a better result than other variations of the R-BERT models and RIFRE. The contributions in this paper are using text augmentation techniques to increase the size of the PERLEX dataset, and preprocessing the PERLEX dataset to fix some of the issues which improves the performance of the latest Persian relation extractor.

\bibliographystyle{aclnatbib}
\bibliography{anthology,acl2021}


\end{document}